# MACHINE-LEARNING BASED METHODOLOGIES FOR 3D X-RAY MEASUREMENT, CHARACTERIZATION AND OPTIMIZATION FOR BURIED STRUCTURES IN ADVANCED IC PACKAGES

Ramanpreet S Pahwa, Soon Wee Ho, Ren Qin, Richard Chang, Oo Zaw Min, Wang Jie, Vempati Srinivasa Rao, Tin Lay Nwe

Agency for Science, Technology and Research (A*STAR)

Singapore

ramanpreet_pahwa@i2r.a-star.edu.sg

Yanjing Yang, Jens Timo Neumann, Ramani Pichumani, Thomas Gregorich

ZEISS

Singapore, Germany, and USA

tom.gregorich@zeiss.com

**ABSTRACT**

For over 40 years lithographic silicon scaling has driven circuit integration and performance improvement in the semiconductor industry. As silicon scaling slows down, the industry is increasingly dependent on IC package technologies to contribute to further circuit integration and performance improvements. This is a paradigm-shift and requires the IC package industry to reduce the size and increase the density of internal interconnects on a scale which has never been done before.

Traditional package characterization and process optimization relies on destructive techniques such as physical cross-sections and delayering to extract data from internal package features. These destructive techniques are not practical with today's advanced packages.

In this paper we will demonstrate how data acquired non-destructively with a 3D X-ray microscope can be enhanced and optimized using machine learning, and can then be used to measure, characterize and optimize the design and production of buried interconnects in advanced IC packages. Test vehicles replicating 2.5D and HBM construction were designed and fabricated, and digital data was extracted from these test vehicles using 3D X-ray and machine learning techniques. The extracted digital data was used to characterize and optimize the design and production of the interconnects and demonstrates a superior alternative to destructive physical analysis.

We report a mAP of 0.96 for 3D object detection, a dice score of 0.92 for 3D segmentation and an average of 2.1um error for 3D metrology on the test dataset.

This paper is the first part of a multi-part report.

**INTRODUCTION**

Machine Learning (ML) and Deep Learning (DL) are fast becoming an integral part of advanced manufacturing and inspection fields such as component design [1], optical inspection [2], and anomaly detection [3]. Such systems, with the availability of vast datasets, have improved significantly over classical handcrafted feature learning approaches in various domains. Some of the improvements made in these domains are also applicable to semiconductor fault inspection and defect detection.

There are various ways to perform defect detection in semiconductor data. One widely used approach is to cross-section faulty chips using SEM and visually inspect these fine-grained images to locate and identify the defects. A more recent approach is to use 3D XRM tools to image the chips without needing to cross-section and then visually identify the buried defects. One key advantage of using XRM non-destructive testing (NDT) techniques is that one can avoid destroying the chips and needing to cross section which is an expensive and time-consuming step.

In this paper, we describe, in detail, how we progressed from fabrication of chips to imaging and 3D metrology using Deep Learning. The paper is structured as follows. In Section II, we will explain the fabrication steps involved in producing HBM and TSV test vehicles. In Section III, we will provide a brief overview of how 3D XRM imaging is performed. Section IV presents the 3D annotation, object detection, segmentation, and metrology steps in detail, highlighting our contributions and observations at each stage. We report our results qualitatively and quantitatively in Section V. Finally, in Section VI we will conclude this paper and discuss future research directions.

## II. FABRICATION

In this section, we describe in detail our approach in the fabrication of test vehicles representing memory and logic dice which utilize fine-pitch interconnects and contain through silicon vias (TSVs).

**Fabrication Process for TSV samples**

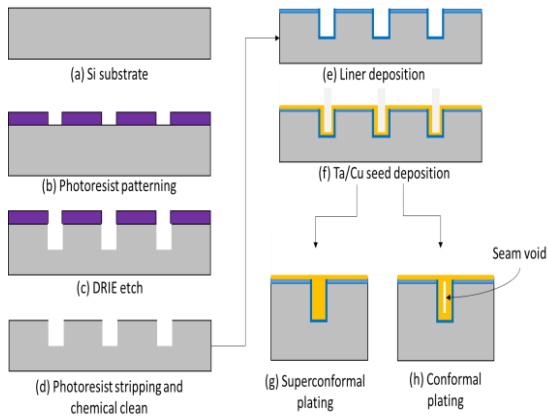

Figure 1. Our approach to fabricate TSVs.

Test vehicles with TSVs were fabricated on 300 mm silicon substrates. A photoresist was spin coated over the Si substrate, followed by mask exposure and then developed to form the via mask patterns. Deep reactive ion etching (DRIE) was used to etch a vertical via in the silicon substrate, through a series of etch and sidewall passivation cycles to achieve the required via depth. The photoresist was removed via a plasma process and a wet chemical cleaning process was used to remove any etch residues in the etched via. A sidewall passivation liner was deposited using plasma enhance chemical vapor deposition (PECVD) process and Ta/Cu seed layer was sputtered as over the via sidewall and field area of the substrate. As shown in Figure 1, a superconformal plating process was used to create TSV samples with no seam voids because the superconformal plating process allows the bottom of the via to be filled at a much greater rate that the sidewall and field area, hence a void free via fill can be achieved. 3D X-ray microscopy was to inspect the samples to detect presence of voids in the plated vias.

**Fabrication Process for Memory and Logic die stack**

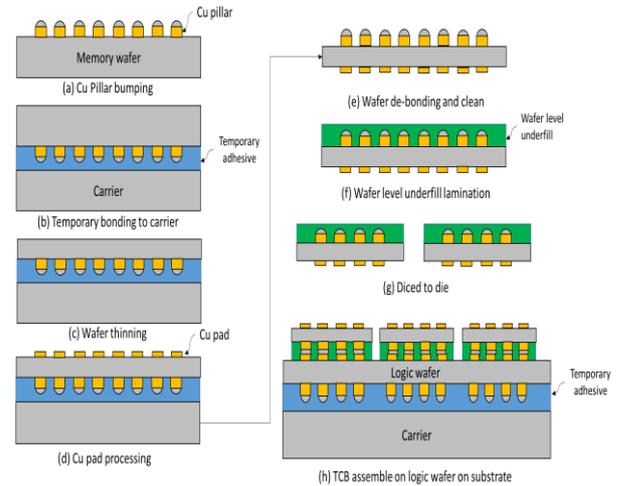

Figure 2. Our approach to fabricate HBMs.

Silicon die were fabricated to represent the memory and logic dies used in a High Bandwidth Memory (HBM) cube. Cu pillar interconnects with SnAg solder caps were fabricated on the memory and logic silicon wafers. The memory die bump diameter and pitch are 20 and 40 micrometers, respectively. As shown in Figure 2, both wafers were temporarily bonded to a carrier after the Cu pillar bumping process and then thinned down to 50 micrometers. Copper pads were plated on the thinned side to function as the receiving pads for the Cu pillar solder bumps. For the memory wafer, the wafer was de-bonded from the carrier after Cu pad processing and a wafer level underfill material was laminated over the Cu pillar bumps. The memory wafer was then diced so the chips could be stacked on the logic wafer. A thermo-compression bonding (TCB) process was used to stack the memory dies on the logic wafer. A 3D X-ray microscopy was then used to non-destructively inspect the stack dies samples to determine the presence of solder non-wetting, voids, and bump to pad alignment from the assembly process.

## III. X-RAY IMAGING AND MEASUREMENTS

Sub-surface, non-destructive imaging data was taken using a ZEISS Xradia 620 Versa 3D X-ray microscope (XRM), which provides a combination of hardware and software tools to perform image reconstruction, image assessment, and 2D/3D data visualization [4]. As shown in Figure 3, the resolution at a distance capability of the

XRM microscope was required in order to obtain the required image resolution[1]. High resolution is achieved by employing a two-stage image formation mechanism which uses multiple objectives to provide both geometric magnification as well as optical magnification. This unique configuration allows high resolution to be maintained even for relatively large samples [5].

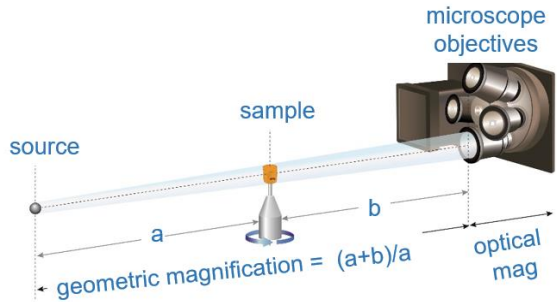

Figure 3. Illustration of the 3D XRM resolution at a distance configuration for 3D CT image acquisition.

Samples were mounted vertically onto standard sample holders and then rotated in incremental steps until a 180-degree rotation was completed for each sample. The X-Ray beam passed through the sample at an incident angle perpendicular to the axis of rotation and is detected by a 2D image sensor. After all of the 2D images from multiple projection were acquired, computation for 3D reconstruction (computed tomography) was performed. This yielded 3D volumetric datasets which were then visualized and processed using a specialized 3D data analytics software tool. This specialized software was used to identify and annotate sub-structures and interconnects to be used in the initial machine learning based training workflow.

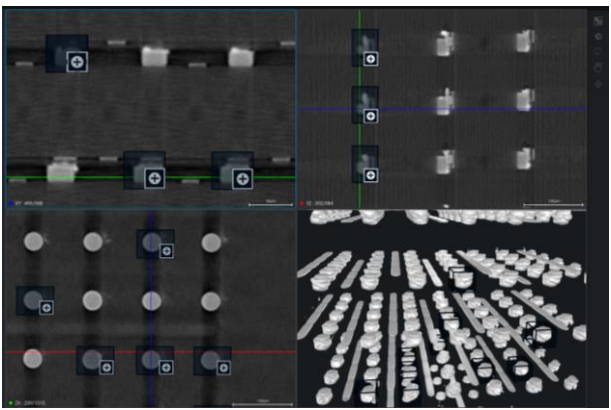

Figure 4. Machine learning-based object detection workflow within the 3D analytics software was used to extract 3D sub-volumes shown above, which are then fed to the segmentation training workflow to build a complete inference model for extracting the desired interconnect features.

---

[1] Resolution-at-a-Distance RaaD™ is a registered trademark of Carl Zeiss Microscopy, Inc.

Figure 4 shows the types of interconnects that were annotated and detected as 3D bounding boxes on 2D virtual slices and overlaid onto 3D volumes as viewed in the analytics tool. The machine learning algorithm utilized by the tool can be trained to detect 3D interconnect structures using a relatively small number of training examples (annotations). Both positive as well as negative examples can be annotated to optimize the accuracy of the object detection results.

## IV. AUTOMATED 3D METROLOGY

There are various approaches to performing 2D/3D metrology. One common way adopted by industry is to use Automated Optical Inspection (AOI) tools for 2D metrology along with Computer Vision techniques [6]. Over the course of time, more sophisticated techniques were developed to account for noise, small variations in image quality, difference in image acquisition, etc. Eventually, these techniques have slowed down to the point where only marginal improvements in accuracy can be achieved. Moreover, most of these techniques might work successfully on one dataset and fail on others.

Recently, Machine learning and Deep learning have made significant strides in improving the performance in object detection and segmentation benchmarks [7, 8]. Using novel neural networks, we can perform pixel and sub-pixel level 2D object detection [9] and image segmentation [3]. Significant effort is being focused in extending these techniques to 3D where data is captured using modern tools such as 3D XRMs, LiDAR, and RGBD cameras [10].

In this section, we describe the process we used to annotate 3D data and to use deep learning and computer vision techniques to perform 3D metrology providing important structural information.

**3D data annotation**

We use a combination of internal tools as well as commonly available open source tools for data annotation. We hand annotated a subset of these components by assigning a class to each 3D region and a tight 3D bounding box was drawn around the region. Thereafter, this region was isolated and for each voxel inside this region, we assign it a foreground or background class depending on the type of segmentation desired. This information was stored offline for model training and evaluation purposes. Table 1 displays our annotation information for object detection and segmentation. Figure 5 shows an example of the 3D annotation of sub-surface structures found in advanced packages.

Table 1. The groundtruth annotation done for training and evaluating our deep-learning based models.

| Component | Object detection | Segmentation |
|---|---|---|
| Structure A | 277 | 13 |
| Structure B | 2023 | 8 |
| Structure C | 1183 | 8 |

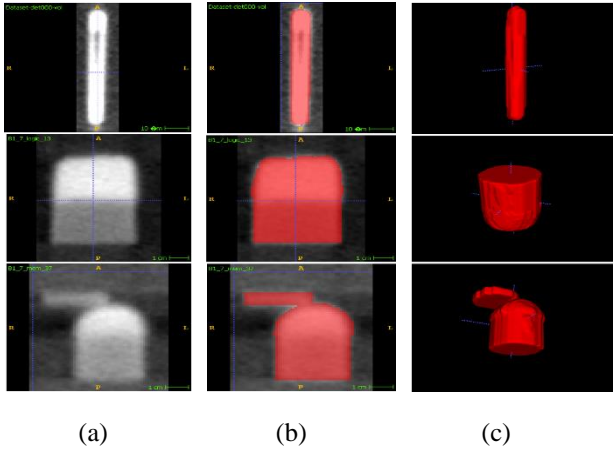

(a)         (b)         (c)

Figure 5. A sample annotation of sub-surface structures found in advanced packages. a) displays the raw 3D data, b) shows an annotated 2D virtual slice and c) shows the full 3D annotation.

**3D object detection**

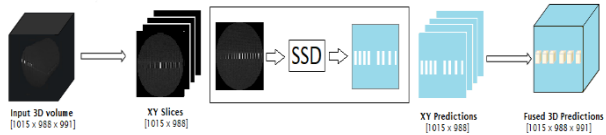

Figure 6. Overall 3D object detection approach. We used a 2D object detector in a slice-and-fuse framework to output 3D bounding boxes.

3D object detection consists of detecting objects in 3D volumes and creating 3D bounding boxes around them (Figure 6). This has been extensively studied and successfully applied to various fields such as autonomous driving or robotics. In this paper, we propose to use a specific slice-and-fuse approach to detect 3D objects in the scans. The idea is to slice the 3D volume into 2D slices, perform 2D object detection on the slices and finally fuse 2D results into 3D bounding boxes [9]. First, this allows us to leverage on the 2D object detection high performance and speed. Second, the computational cost and memory requirement also both increase cubically with voxel resolution. Thus, it is infeasible to train a voxel-based 3D detection model with high resolution inputs such as large 3D scans [11]. Figure 6 shows our overall approach.

Convolutional neural networks have demonstrated excellent results on 2D images object detection. Many successful networks have been published such as SSD [12], YOLO [13], and Faster-RCNN [14]. Our choice of network was based on the need for highest throughput, the ability to extract significant features in the image, and the ability to predict bounding boxes with a confidence score. Each 3D volume scan ($N_x$ x $N_y$ x $N_z$) is sliced along the z-direction giving $N_z$ images of size ($N_x$ x $N_y$).

The network takes the images $I_{i=1,...,N_z}$ as input and outputs a list of bounding boxes defined by $[x_{min}, x_{max}, y_{min}, y_{max}]_k$ corresponding to the $k$ detected objects. Then, the bounding boxes from all $N_z$ images are concatenated into 3D bounding boxes defined by [x, y, z, w, h, d] where [x, y, z] represents the coordinates of the starting point of the cuboid and w, h, d its dimensions.

Our 2D object detector network is pretrained on COCO dataset [15]. We used our own training data which consisted of annotated images of sub-surface structures found in advanced packages for fine-tuning the model. We trained 3 independent models in this paper, one for detecting TSV bumps, one for detecting memory bumps and one for detecting logic bumps. The implementation details are as follows. We trained with an optimizer running on a GeForce RTX 2080 Ti GPU accelerator.

**3D binary segmentation**

The above-mentioned 3D object detection module generates individual 3D ROIs for different type of components. We segmented foreground object from the background for each ROI using 3D Anisotropic Convolutional Neural Network (CNN) [7] for segmentation. The network takes a stack of input 2D slices with a large receptive field and a relatively small receptive field in the orthogonal out-plane direction. As the network has a small receptive field in the out-plane direction compared to large receptive field in 2D plane, the network is referred to as Anisotropic CNN [7].

The network involves multiple residual blocks with anisotropic convolution, dilated convolution, and multi-scale prediction [8]. The convolution layers are followed by batch normalization layers and activation layers. The activation layers use Parametric Rectified Linear Unit (PReLU) which provide better segmentation performance than traditional rectified units [8]. The entire network has minimal down-sampling layers to avoid large image resolution reduction and loss of segmentation details. The residual blocks of the network are to create identity mapping connections and to bypass parameterized layers of the network. The input of each residual block is directly added to the output and hence, each residual block can learn residual functions with reference to the input. This helps the network for faster convergence during training.

The network learns local (low-level) features in shallow layers and global (high-level) features in deep layers. For effectiveness in segmentation, both low-level and high-level features are combined. Hence, multiple intermediate predictions are obtained at different depths of the network and a concatenation of these predictions are fed to the last convolution layer to get the final score map.

**3D metrology**

We performed computer vision based 3D metrology after obtaining binary segmentation of the 3 components.

For each of these components, the data was loaded per 2D binary mask in sequential order. Thereafter, for each mask, we applied a series of morphological functions to remove any outlier foreground regions that might have been present in these binary 2D masks. After removing noise, as each mask contained one foreground object, we estimated the best contour (2D boundaries) for the given binary mask. We computed a tight-fitting rectangle around these boundaries using OpenCV [16]. This rectangle was represented by a vector - [x, y, w, h] where *x, y* represents the starting point of rectangle and *w, h* represents the width and height of rectangle. We processed each mask sequentially to obtain the boundary information for them.

Thereafter, we processed the rectangle information together and compute the starting point when we started seeing the regions and would stop when foreground regions disappeared. This provided us the important depth information – where the 3D ROIs started and ended in a 3D scan. This step allowed us to compute the tight-fitting 3D location of these ROIs accurately as a cuboid. Each ROI is represented by [x, y, z, w, h, d] where *x, y, z* represents the 3D starting point of a cuboid and *w, h, d* represent the width, height, and depth of the cuboid.

**V. RESULTS**

In this section, we report our results for 3 steps – 3D Object detection, 3D segmentation, and 3D metrology. For evaluation, we divided our data into 3 sets: one each for training, validation and testing as shown in Table 2.

Table 2. The division of annotated data into train, validation, and test subset for our experiments.

| Components | 3D detection | | | 3D segmentation | | |
|---|---|---|---|---|---|---|
| | Train | Val | Test | Train | Val | Test |
| Structure A | 221 | 0 | 56 | 7 | 2 | 4 |
| Structure B | 1014 | 0 | 169 | 5 | 1 | 2 |
| Structure C | 1734 | 0 | 289 | 5 | 1 | 2 |

**3D object detection**

The 3D object detection results for the test vehicles are shown in Table 3. It includes 2D and 3D evaluation using the Intersection-Over-Union (IoU) metric explained in Figure 7. This is used to measure the accuracy of an object detector and is computed by the ratio between the area of overlap and area of union between the predicted bounding box and the ground truth bounding box.

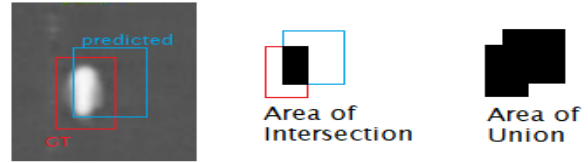

Figure 7. The Intersection-Over-Union metric is computed using the predicted and Ground Truth bounding boxes. It is the ratio of the area of intersection and the area of union between them.

The same principle is then applied to estimate the 3D IOU on 3D bounding boxes instead of 2D. Once the ratio is computed, we used a 0.5 threshold to determine whether the prediction is correct (True Positive) or incorrect (False Positive). The precision of our detector is then defined by:

$$Precision = \frac{TP}{TP + FP}$$

By averaging the precision values on all the predicted bounding boxes, we obtained the mean Average Precision reported in Table 3. Since the 3D mAP would be 1.0 because all structures are detected, we included the raw 3D IOU average value instead. This metric shows the accuracy of the 3D detector. Figure 8 displays our 3D detection results for three types of sub-surface structures.

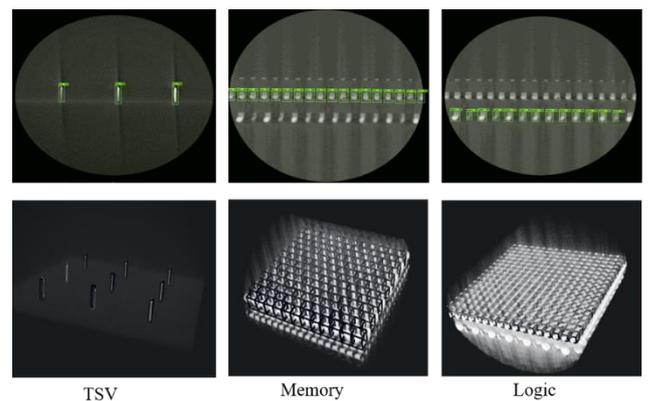

Figure 8. 2D and 3D detection results for Structures A, B, and C. The 2D bounding boxes from all slices are concatenated to generate 3D bounding boxes.

Table 3. Test Results for our 3D object detection. We display our mean 2D IoU and 3D IoU for each test structure.

| Test Structure | 2D mAP @ IoU = 0.5 | 3D IoU |
|---|---|---|
| Structure A | 0.955 | 0.697 |
| Structure B | 0.999 | 0.686 |
| Structure C | 0.986 | 0.610 |

**3D binary segmentation**

Each 3D ROI has 3 views: axial, sagittal, and coronal. We trained an anisotropic CNN model for each of the 3 views for binary segmentation (foreground vs. background). We then fused the output of CNNs in three orthogonal views for robust voxel segmentation. During testing, predictions from models of three views are averaged to calculate the segmentation accuracy in terms of Dice score. The Dice score is expressed as:

$$Dice(X,Y) = \frac{2 \times |X \cap Y|}{|X| + |Y|}$$

where X and Y represent the output binary mask and groundtruth binary mask respectively. Dice score of '1' means perfect automatic segmentation and '0' means segmentation is completely wrong. We use a patch size of 19x128x128 as an input to our network.

Table 4. 3D dice score for binary segmentation on test data

| Structure | 3D dice score |
|---|---|
| Structure A1 | 0.9158 |
| Structure A2 | 0.9127 |
| Structure A3 | 0.9121 |
| Structure A4 | 0.9317 |
| Structure B1 | 0.9666 |
| Structure B2 | 0.9581 |
| Structure C1 | 0.9286 |
| Structure C2 | 0.9292 |

Dice scores for each of the 3D ROIs in test set are listed in Table 4. The table shows promising dice scores for all ROIs of sub-surface structures. Although we used very little data to train our models, we achieved above 90% segmentation accuracies for all 3D ROIs. Our visual results are shown in Figure 9.

**3D metrology**

We pre-processed the segmented data by performing data cleaning, noise removal, and then used OpenCV based metrology to estimate 3D cuboids around the three structures. The results of our estimates and ground truth estimates (measured using 3D XRM visualization tools) are reported in Table 5. Our 3D metrology results are close to actual ground truth results with an MSE of 2.1 um. Currently, we use only 6 samples per component to train our models for object detection and segmentation. With more annotated data, we expect to further improve our models and results accuracy. Table 5 shows immense promise in our approach for estimating dimensions accurately for various components present in die stacking process.

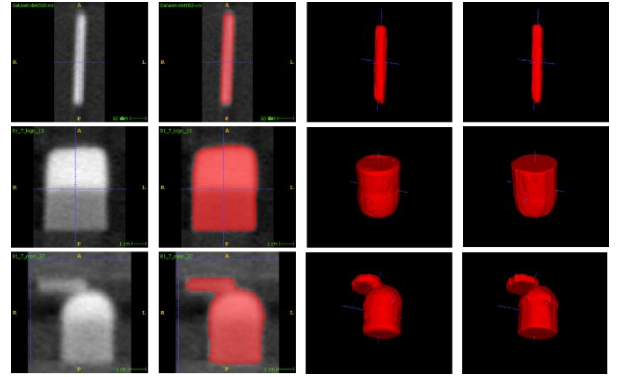

Figure 9. We illustrate the visual segmentation results of our approach for sub-surface structures A, B, and C in the three rows, respectively. First column displays the raw data. Second and third column shows our segmentation on a sample 2D virtual slice and our full 3D segmentation. Last column shows the groundtruth 3D annotation. Our segmentation results are extremely close to the groundtruth annotation.

Table 5. Automated 3D metrology and groundtruth measurements on test data (in μm). Results are reported in um. W, H, D correspond to the 3 dimensions – width, height, and depth

| Structure | Our metrology | | | Groundtruth | | |
|---|---|---|---|---|---|---|
| | W | H | D | W | H | D |
| Structure A1 | 8.4 | 65.1 | 8.4 | 8.4 | 61.6 | 9.8 |
| Structure A2 | 9.1 | 64.4 | 8.4 | 7.7 | 61.6 | 9.1 |
| Structure A3 | 9.1 | 60.9 | 7.7 | 8.4 | 60.2 | 7.0 |
| Structure A4 | 9.8 | 59.5 | 7.7 | 8.4 | 58.1 | 7.7 |
| Structure B1 | 32.9 | 43.4 | 32.9 | 33.6 | 42.0 | 32.2 |
| Structure B2 | 32.2 | 43.4 | 32.9 | 34.3 | 42.7 | 33.6 |
| Structure C1 | 33.6 | 35.7 | 25.2 | 32.9 | 35.0 | 25.9 |
| Structure C2 | 32.9 | 35.7 | 25.2 | 32.9 | 33.6 | 25.2 |

**VI. CONCLUSION**

We presented our approach on automated 3D metrology for various test vehicles replicating 2.5D and HBM construction in this work. First, we demonstrated how test vehicles replicating 2.5D and HBM construction were designed and fabricated. We also described how 3D digital data could be acquired and extracted non-destructively using a 3D X-ray microscope. This data was then annotated using sophisticated tools and used to train deep learning models for object detection and binary segmentation. We then used our trained models to identify other test vehicles and perform 3D metrology to test their performance. Our results show promising results for automated 3D metrology. We expect our results to further improve with more training data.

In the future, we will extend our approach to more test vehicles and expand the capability of the system to include attribute and anomaly detection. We also intend to continue working to incorporate ~~include~~ multi-class segmentation so that sub-surface structures can be better sub-classified into base materials and compare our results with state-of-the-art tools in future.

ACKNOWLEDGEMENTS

The authors would like to acknowledge the support of Center of Excellence set up between I$^2$R, IME, and Zeiss in Singapore for supporting this study. This work is supported by Economic Development Board (EDB), Singapore the under its IAF-ICP Grant no. I1901E0048 and administered by the Agency for Science, Technology and Research

REFERENCES

[1] J. Wang, Y. Ma, L. Zhang, R. X. Gao, and D. Wu, "Deep learning for smart manufacturing: Methods and applications," in Journal of Manufacturing Systems, vol. 48, pp. 144–156, 2018

[2] D. Weimer, B. Scholz-Reiter, and M. Shpitalni, "Design of deep convolutional neural network architectures for automated feature extraction in industrial inspection," in CIRP Annals, vol. 65, no. 1, pp. 417–420, 2016.

[3] R. S. Pahwa *et al*., "FaultNet: Faulty Rail-Valves Detection using Deep Learning and Computer Vision," in IEEE Intelligent Transportation Systems Conference (ITSC), Auckland, New Zealand, 2019, pp. 559-566, doi: 10.1109/ITSC.2019.8917062.

[4] T. Gregorich, M. Terada, C. Hartfield, A. Gu and J. Vardaman, "Non-destructive 3D Characterization of Application Processor Panel Level Package Used in Galaxy Smartwatch", IWLPC, 2019

[5] M.Z. Syahirah, B. Zee, T. Gregorich, A. Gu, Y. Yang W. Lee, "High-resolution 3D X-ray Microscopy for Structural Inspection and Measurement of Semiconductor Package Interconnects", IPFA, 2019

[6] W. Osten, ed. Optical inspection of Microsystems. CRC press, 2019.

[7] G. Wang, W. Li, S. Ourselin, and T. Vercauteren, "Automatic brain tumor segmentation using cascaded anisotropic convolutional neural networks," in International MICCAI brain-lesion workshop. Springer, 2017, pp. 178–190.

[8] T. L. Nwe, Z. M. Oo, S. Gopalakrishnan, D. Lin, S. Prasad, S. Dong, Y. Li, R. S. Pahwa, "Improving 3d Brain Tumor Segmentation With Predict-Refine Mechanism Using Saliency And Feature Maps", in IEEE International Conference on Image Processing, ICIP 2020.

[9] A. Yang, "3D Object Detection from CT Scans using a Slice-and-fuse Approach," Doctoral dissertation, Robotics Institute, 2019.

[10] R. S. Pahwa, J. Lu, N. Jiang, T. T. Ng and M. N. Do, "Locating 3D Object Proposals: A Depth-Based Online Approach," in *IEEE Transactions on Circuits and Systems for Video Technology*, vol. 28, no. 3, pp. 626-639, March 2018, doi: 10.1109/TCSVT.2016.2616143.

[11] Z. Liu and H. Tang and Y. Lin and S. Han, "Point-Voxel CNN for Efficient 3D Deep Learning," In Advances in Neural Information Processing Systems. 2019.

[12] W. Liu, D. Anguelov, D. Erhan, C. Szegedy, S. Reed, C-Y. Fu, and A. Berg, "SSD: Single Shot MultiBox Detector," in European conference on computer vision. Springer, Cham, 2016

[13] J. Redmon, S. Divvala, R. Girshick, and A. Farhadi, "You only look once: Unified, real-time object detection," in Proceedings of the IEEE conference on computer vision and pattern recognition, 2016.

[14] S. Ren, K. He, R. Girshick, and J. Sun, "Faster R-CNN: Towards Real-Time Object Detection with Region Proposal Networks," in Advances in neural information processing systems, 2015.

[15] T. Y. Lin, ... C. L. Zitnick, "Microsoft coco: Common objects in context," in European conference on computer vision 2014.

[16] G. Bradski, A. Kaehler, "Learning OpenCV: Computer vision with the OpenCV library", O'Reilly Media, Inc., 2008.